\documentclass[letterpaper]{article} 
\usepackage{aaai2026}  
\usepackage{times}  
\usepackage{helvet}  
\usepackage{courier}  
\usepackage[hyphens]{url}  
\usepackage{graphicx} 
\usepackage{natbib}  
\usepackage{caption} 
\usepackage{algorithm}
\usepackage{algorithmic}
\usepackage{newfloat}
\usepackage{listings}
\usepackage{amsmath, amsfonts, amssymb}
\usepackage{multirow}
\usepackage{booktabs}
\usepackage{array}
\usepackage{caption}
\usepackage{graphicx}
\usepackage{siunitx}
\usepackage{makecell}
\usepackage{float}
\usepackage{pdfpages}
\usepackage{xcolor}
\definecolor{deepblue}{RGB}{0,0,128}

\DeclareCaptionStyle{ruled}{labelfont=normalfont,labelsep=colon,strut=off} 
\floatstyle{ruled}
\newfloat{listing}{tb}{lst}{}
\floatname{listing}{Listing}

\title{UNeMo: Collaborative Visual-Language Reasoning and Navigation via a Multimodal World Model}
\author{
    Changxin Huang\textsuperscript{\rm 1, \rm 2}, Lv Tang\textsuperscript{\rm 1, \rm 2}, Zhaohuan Zhan\textsuperscript{\rm 3}, Lisha Yu\textsuperscript{\rm 4}, Runhao Zeng\textsuperscript{\rm 5}, \\
    Zun Liu\textsuperscript{\rm 1, \rm 2},
    Zhengjie Wang\textsuperscript{\rm 6}, Jianqiang Li\textsuperscript{\rm 1, \rm 2}\thanks{Corresponding author}
}
\affiliations{
    \textsuperscript{\rm 1} School of Artificial Intelligence, Shenzhen University\\
    \textsuperscript{\rm 2} National Engineering Laboratory for Big Data System Computing Technology, Shenzhen University\\
    \textsuperscript{\rm 3} Department of Engineering, Shenzhen MSU-BIT University\\
    \textsuperscript{\rm 4} School of Computer Science and Engineering, Sun Yat-sen University\\
    \textsuperscript{\rm 5} Artificial Intelligence Research Institute, Shenzhen MSU-BIT University\\
    \textsuperscript{\rm 6} School of Mechatronical Engineering, Beijing Institute of Technology\\
    {}
    huangchx@szu.edu.cn, lijq@szu.edu.cn
}

\usepackage{bibentry}

\begin{document}
\maketitle
\begin{abstract}
Vision-and-Language Navigation (VLN) requires agents to autonomously navigate complex environments via visual images and natural language instructions—remains highly challenging. Recent research on enhancing language-guided navigation reasoning using pre-trained large language models (LLMs) has shown promising prospects. However, the reasoning of such methods is limited to the linguistic modality, lacking visual reasoning capabilities. Moreover, existing reasoning modules are optimized separately from navigation policies, leading to incompatibility and potential conflicts in optimization objectives.
To tackle these challenges, we introduce UNeMo, a novel framework designed for the collaborative optimization of visual state reasoning and navigational decision-making. It introduces a Multimodal World Model (MWM) that takes visual features, language instructions, and navigational actions as inputs to jointly predict subsequent visual states, enabling cross-modal reasoning. Via a Hierarchical Prediction-Feedback (HPN) mechanism, MWM collaborates with navigation policies: the first layer generates actions using current vision-and-language features; MWM then infers post-action visual states to guide the second layer’s fine-grained decisions. This forms a dynamic bidirectional promotion mechanism where MWM reasoning optimizes navigation policies, while policy decisions feedback to improve MWM’s reasoning accuracy. Experiments on R2R and REVERIE datasets show UNeMo outperforms state-of-the-art methods by 2.1\% and 0.7\% in navigation accuracy for unseen scenes, validating its effectiveness.
\end{abstract}

\section{Introduction}
Vision-and-Language Navigation (VLN) requires an embodied agent to achieve autonomous navigation from the current to the target position in an unknown environment based on visual observations and language instructions \cite{anderson2018vision_r2r, krantz2020beyond_vlnce, qi2020reverie}. This capability constitutes the core foundation of an embodied intelligent system. This task not only needs to address the problem of efficient fusion of cross-modal information \cite{gu2022vision_svy, wu2024vision_svy} but also poses new challenges to the construction of a perception-reasoning collaborative mechanism in dynamic environments \cite{zhang2024vision_svy}.

Traditional VLN methods use tailored end-to-end deep learning frameworks, such as sequence-to-sequence models \cite{nips/FriedHCRAMBSKD18}, attention-based cross-modal fusion \cite{cvpr/Hong0QOG21}, and graph-based representation learning \cite{eccv/KrantzWMBL20}. These jointly optimize vision-language alignment and action policy learning to map multimodal inputs to navigation decisions \cite{an2024etpnav, yue2024safe}. While performing well on large task-specific datasets, they show poor policy generalization in out-of-distribution environments or with novel instructions.

\begin{figure}[t]
\centering
\includegraphics[width=0.98\columnwidth]{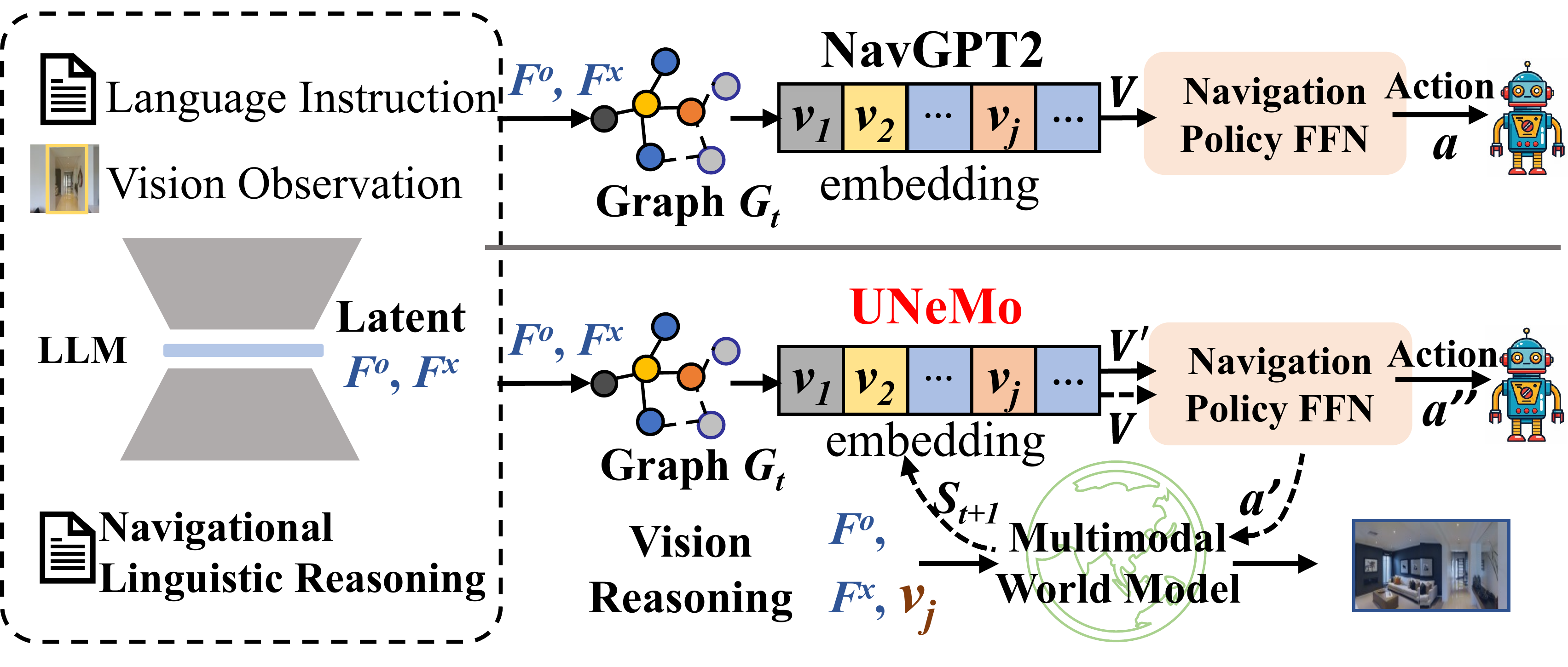}
\caption{Comparison of main differences between UNeMo and NavGPT2. UNeMo introduces MWM for visual state reasoning and achieves joint optimization with the navigation policy via a hierarchical predictive-feedback navigator.}
\label{fig:introduction overview}
\end{figure}

Large language models (LLMs) have recently shown strong abilities in knowledge generalization and complex reasoning, prompting research into their use for enhancing VLN with promising results. Researchers have attempted to use LLMs to assist VLN in tasks such as navigation instruction parsing \cite{yang2024llm}, enhancing scene descriptions \cite{wu2024voronav}, and extracting task knowledge \cite{shah2023lm, schumann2024velma, wang2023voyager}. NavGPT further leverages the decision-making capabilities of LLMs to directly predict navigation actions \cite{aaai/ZhouHW24}. However, due to the lack of training on navigation data, the performance of such methods is inferior to traditional VLN-specialized approaches. 

Accordingly, NavGPT2 employs navigation data to conduct knowledge distillation on LLMs, thereby constructing a navigation-specialized LLM \cite{zhou2024navgpt2}, as illustrated in Fig. \ref{fig:introduction overview}. Equipped with a navigational reasoning LLM, this approach achieves performance on par with VLN-specialized methods. However, NavGPT2 is trained exclusively via language-guided reasoning modules, rendering it incapable of visual state reasoning. Furthermore, its navigational reasoning module is frozen during the training phase of navigation policies, creating a disconnect between the reasoning process and navigation optimization. This hinders the dynamic refinement of navigational reasoning capabilities in response to policy training progress, thereby imposing inherent limitations on navigation performance.

To address the aforementioned issues, we propose an \underline{U}nlock \underline{Ne}xt \underline{Mo}ment (UNeMo) framework, which enables the collaborative optimization of cross-modal reasoning and navigation policies. Specifically, the framework introduces a Multimodal World Model (MWM) to implement vision state reasoning: it takes the agent’s current observed visual features, language instruction features, and navigation action prediction results as inputs, and predicts the next visual state through joint modeling, as shown at the bottom of Fig. \ref{fig:introduction overview}. The MWM and navigation policies adopt a synchronous optimization mechanism, endowing the navigation policies with vision-language cross-modal reasoning capabilities.
To enhance the collaborative effectiveness between the world model and navigation policies, this study designs a Hierarchical Prediction-Feedback Navigator (HPFN). This navigator employs a hierarchical decision-making architecture: first, it performs coarse-grained navigation action prediction based on current vision-language features; subsequently, it uses the MWM to predict the visual state of the future node that the agent will reach after executing the action, and integrates this visual feedback feature into the decision-making process of the navigation policy to guide fine-grained navigation action prediction.

Through the synchronous optimization of multimodal reasoning and navigation decision-making, UNeMo constructs a dynamic bidirectional promotion mechanism: the decisions of the navigation policy feed back into the state prediction of the world model, while the reasoning results of the world model continuously optimize the navigation policy. 

The contributions can be summarized as follows:
\begin{itemize}
    \item We propose MWM, a multimodal world model that jointly captures visual observations, language instructions, and navigation-action predictions to enable dynamic visual state reasoning.
    \item We design a hierarchical prediction-feedback navigator, employing a closed‑loop interaction paradigm of coarse‑grained action prediction → world‑model visual state feedback → fine‑grained decision refinement, thereby strengthening the synergy between navigational reasoning and decision making.
    \item Extensive experiments on the R2R and REVERIE benchmarks demonstrate that, in unseen environments, UNeMo outperforms state‑of‑the‑art methods by 2.1\% and 0.7\% in success rate, respectively, underscoring its efficacy in enhancing generalization and navigation accuracy in complex VLN scenarios.
\end{itemize}

\section{Related Work}
Traditional research in Vision-Language Navigation (VLN) has centered on state representation learning, including recurrent neural network (RNN)-based historical state modeling ~\cite{Hong_2021}, sequential encoding for temporal information processing ~\cite{chen2021history}, topological map construction for spatial connectivity ~\cite{an2024etpnav, chen2022think_topo}, and grid map development for precise localization ~\cite{liu2023bird_map}, Additionally, frameworks such as SUSA ~\cite{zhang2024agent_SUSA} have introduced hybrid semantic-spatial representations to enhance vision-language alignment. Recent advances have shifted toward leveraging Large Language Models (LLMs) to enhance navigation capabilities.

\textbf{Navigation with LLM.} LLMs, with their strong semantic understanding and reasoning abilities, have been increasingly applied to navigation agents~\cite{huang2022inner, yao2023react, shah2023lm, schumann2024velma, wang2023voyager}. Early approaches decoupled LLMs from core navigation systems—using GPT-2 for subtask hints~\cite{qiao2023march} or GPT-4 prompts to translate instructions into actions~\cite{song2025towards}—limiting holistic integration. Recent work embeds LLMs directly, such as zero-shot NavGPT~\cite{aaai/ZhouHW24}, MapGPT~\cite{chen2024mapgpt}, MC-GPT~\cite{zhan2024mc}, and fine-tuned LLaMA-7B variants like LangNav~\cite{pan2023langnav} and NavCoT~\cite{lin2025navcot}. Despite improvements, these agents still lag behind specialized VLN models in accuracy and transparency. NavGPT-2, which freezes the LLM as a vision-language encoder with imitation learning~\cite{zhou2024navgpt2}, lacks dynamic scene adaptation. In contrast, our work integrates a future-state prediction module to dynamically forecast environmental changes for enhanced decision-making.

\textbf{World Model for VLN.} World models~\cite{Ha_Schmidhuber_2018} compress high-dimensional sensory data into dynamic state representations, supporting embodied decision-making. Prior studies explored diverse architectures for environmental feature encoding ~\cite{Chang_2020, Purushwalkam_2020}, while sampling-based planners used these representations for self-supervised policy learning~\cite{Sekar_2020, Chen20a}. Pathdreamer~\cite{Koh_2021_ICCV} and Dreamwalker~\cite{wang2023dreamwalker} predict future observations (pixel-level or abstract) to extend planning horizons but focus solely on forward exploration. LookAhead ~\cite{wang2024lookahead} employs a hierarchical neural radiance field to generate multi-scale future embeddings. However, these methods separate state prediction from policy learning. Our work introduces a joint framework for simultaneous future state prediction and proposes a hierarchical prediction-feedback mechanism that co-optimizes state inference and action selection, fully integrating predicted future states into navigation decisions for more precise navigation.

\begin{figure*}[t]
\centering
\includegraphics[width=\textwidth]{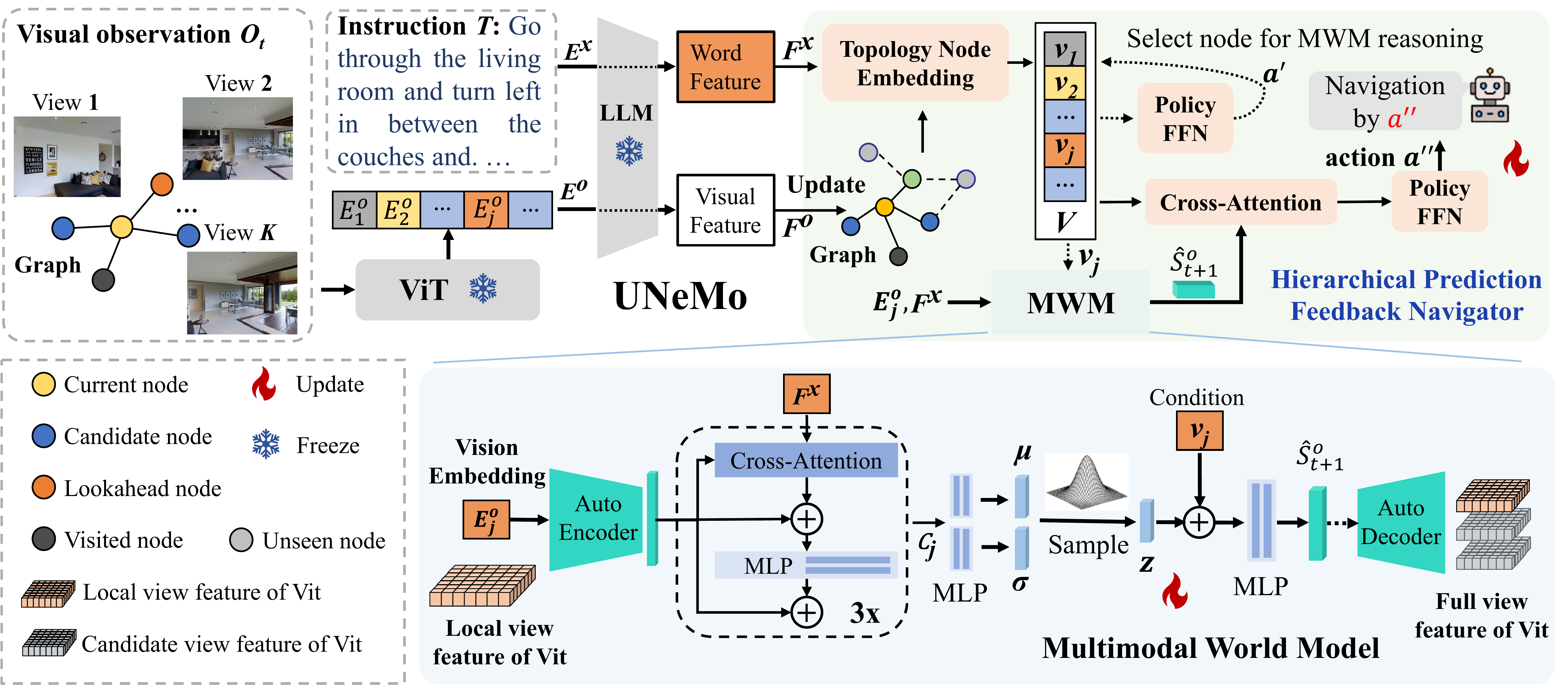}
\caption{Overview of the UNeMo framework. Similar to NavGPT2, the input visual observation $O_t$ and instruction Tare processed through a pre-trained LLM encoder to obtain visual and word features, respectively. UNeMo introduces two key components: (1) The Multimodal State Reasoning module receives partial-view representations of the highest-scoring action node and predicts its complete post-execution visual state; (2) The Decision Making module fuses these state-reasoning results with linguistic features for final navigation action selection.}
\label{fig:framework}
\end{figure*}

\section{Preliminary}
\subsection{Problem setup}

In the VLN task, agents need to parse multimodal information to achieve goal-oriented path planning in discrete topological spaces. Specifically, an agent must understand the global natural language instruction $I$ and select a navigational action $a_t$ based on the current visual observation ${v}_t$. This process is repeated until the target location is reached or the number of steps exceeds a threshold.

The action prediction in VLN can be formulated as a node-selection problem. Early methods confined this space to the local neighborhood of the agent’s current node; more recent approaches have extended it to the global scale by constructing an explicit topological map \cite{chen2022think_topo, zhou2024navgpt2}. The agent builds a topological map $\mathcal{G} = (\mathcal{V}, \mathcal{E})$ based on the nodes it has visited and the nodes within its visual range, where $\mathcal{V}=\{v_1,v_2,...,v_N\}$ denotes the set of N navigable nodes, and $\mathcal{E}$ represents the set of connecting edges between nodes. At each time step $t$, the agent acquires observations of adjacent navigable nodes at the current position and updates the topological map $\mathcal{G}_t$. For visited nodes, the node embedding is the observation at that node. For unvisited but partially observable nodes, the node embedding aggregates the average of all partial-view embeddings observed from other visited nodes. The final action prediction $a_t$ corresponds to selecting a node $v \in \mathcal{V}$. 

\subsection{LLM-based navigator}
We adopt NavGPT2 \cite{zhou2024navgpt2}, one of the most advanced LLM-based navigation approaches to date, as our base model. NavGPT2 incorporates three core components: node embedding, decision making, and loss function. For node embedding, it employs a Topology Node Embedding (TNE) encoder. Specifically, let $E^o$ and $E^x$ denote the visual (observation) and textual (instruction) features, respectively. These are jointly encoded by a pretrained LLM to produce the corresponding features $F^o$ and $F^x$. The TNE encoder then projects these embeddings into a shared topological space to form the node embeddings used for navigation:
\begin{equation}
\label{eq:tne}
\mathcal{V}_t = \text{TNE}(F^o, F^x).
\end{equation}

For decision making, a two-layer feedforward network is used to predict the score of each navigational action, and the node with the highest score is selected for action execution. The decision-making process is as follows:
\begin{equation}
\label{eq:navgpt2_action}
a_t = \mathop{\max}\big(\text{Softmax}(\text{FFN}(\mathcal{V}_t) )\big).    
\end{equation}

For the loss function, NavGPT2 integrates behavioral cloning and DAgger \cite{ross2011reduction} objectives into a single training criterion:
\begin{equation}
\label{eq:NavGPT2_loss}
\mathcal{L}_{\text{base}}=\lambda \mathcal{L}_{\text{BC}} + \mathcal{L}_{\text{DAG}},
\end{equation}
where $\mathcal{L}_{\text{BC}} = -\sum_{t=1}^{T} \log \pi(v_{t}^{*} \mid \mathcal{I}, \mathcal{G}_{t})$ optimizes for expert demonstrations $v_{t}^{*}$, while $\mathcal{L}_{\text{DAG}} = -\sum_{t=1}^{T} \log \pi(\tilde{v}_{t}^{*} \mid \mathcal{I}, \tilde{\mathcal{G}}_{t})$ leverages pseudo-labels $\tilde{v}_{t}^{*}$ derived from shortest-path analysis on the agent's topological map $\tilde{\mathcal{G}}_t$, with hyperparameter $\lambda$ balancing their contributions.

\section{Method}
\subsection{Overview}
Our framework is illustrated in Figure \ref{fig:framework}. Similar to NavGPT2, input visual observations \( O_t \) and instructions \( T \) are processed via a pre-trained LLM encoder to generate visual and word features, respectively. To endow the agent with visual reasoning capabilities, we introduce a Multimodal World Model (MWM) for visual state prediction: this module takes the current visual state encoding, instruction encoding, and navigation action as inputs, and predicts the next visual state the agent will encounter. This predicted visual state updates the embedding of each navigation node, enriching the agent’s representation with forward-looking information to support more informed navigation decisions. To mitigate objective conflicts arising from separate optimization of the navigation policy and reasoning module, we further propose a Hierarchical Prediction-Feedback Navigator. This mechanism enables bidirectional interaction: navigation policy decisions feed back to refine the world model’s state predictions, while the world model’s reasoning outputs continuously optimize the navigation policy. Together, these components facilitate the joint optimization of navigation reasoning and decision-making processes.

\subsection{Multimodal World Model}
While topological maps encode historical states—encompassing visual observations and node connectivity—they lack information about future states. Given that future state prediction enhances navigation performance~\cite{li2023improving, wang2023dreamwalker, wang2024lookahead}, we introduce a Multimodal World Model (MWM) to predict future visual states, leveraging the topological map's representation of past states. We employ Conditional Variational Autoencoders (CVAE) \cite{NIPS2015_cvae} to construct the MWM, which comprises an encoder for aligning and fusing multimodal inputs, and a decoder for reasoning about the next-step visual state.

The encoder uses cross-attention to align visual observations with language instructions. At time $t$, the navigator generates a node-scoring distribution $a_t$ over the topological map, selecting the highest-scoring node $j$. The partially observed view embedding $E_{j}^o \in \mathbb{R}^{1 \times 768}$ (visual observation of neighbor $j$) is combined with instruction feature $F^x \in \mathbb{R}^{N \times 768}$; both feed into cross-attention for fusion, yielding ${c}_{j} \in \mathbb{R}^{ 1 \times 768}$, formally:
\begin{equation}
\begin{aligned}
\label{eq:MWM_Encoder_Res_Connection}
c_{j} = \text{Softmax}\Bigg( \frac{\mathcal{W}_qE^o_j (\mathcal{W}_kF^x)^\top}{\sqrt{d}} \Bigg) \mathcal{W}_vF^x,
 \end{aligned}
\end{equation}
$\mathcal{W}_*$ denotes learnable projection matrices, with subscripts $q,k,v$ indicating mappings to attention query, key, and value vectors; $d$ is the scaling factor. Specifically, observed-view embedding $E_{j}^o$ projects to the query, and instruction embedding $F^x$ projects to key/value. Stacking $L=3$ cross-attention layers yields the final cross-modal embedding ${c}_{j} \in \mathbb{R}^{1 \times 768}$ (matching query dimensionality). This fused feature is fed through two MLPs to estimate future-state distribution parameters: mean $\mu_t$ and variance $\sigma_t^2$.

In the decoder, the mean $\mu_t$ and variance $\sigma_t^2$ are first reparameterized to sample the latent variable $z_t = \mu + \sigma \odot \varepsilon, \varepsilon \sim \mathcal{N}(0, 1)$,
In the decoder, $\mu_t$ and $\sigma_t^2$ are reparameterized to sample latent $z_t = \mu + \sigma \odot \varepsilon$ ($\varepsilon \sim \mathcal{N}(0, 1)$). Node $j$'s intrinsic embedding $v_j^t$ is concatenated with $z_t$ as a conditional feature; an MLP then uses this vector to predict future visual-state embedding $\hat S_{t+1}^o = \mathrm{MLP}(z_t; {v}_{j}^t)$. Thereforce, The entire visual state reasoning process of MWM can be formulated as:
\begin{equation}
\label{eq:MWM_Decoder}
\hat S_{t+1}^o = \mathrm{CVAE}(E^o_j, F^x|v_j), 
\end{equation}
this reasoning visual-state is integrated into the subsequent Hierarchical Prediction-Feedback Navigator to enhance navigation performance.

\subsection{Hierarchical Prediction-Feedback Navigator}
Although the MWM enables prediction of future visual states, integrating these predictions into the navigation policy remains an open challenge. Here we employ the prediction of MWM to upadate node embeddings and guide decision making.

To update the node embeddings, we first pass topological‐map node embeddings $\mathcal{V}_t = \{{v}_{t,i}\}_{i=1}^K$  through a two‐layer feedforward network to predict the look ahead action $a_t^{\prime}$. Then we employ the MWM to predict the visual-state embedding after executing the lookahead exploration $\hat S_{t+1}^o$ as described in Eq. \ref{eq:MWM_Decoder}. This prediction $\hat S_{t+1}^o$ embodies the expected observation once the agent arrives at node $j$. We introduce a cross-attention module to enable deep interaction between the topological-map node features and the MWM’s predicted state. Specifically, we treat the node embeddings $\mathcal{V}_t$ as query vectors, and the predicted next-state embedding $\hat S_{t+1}^o$ as key/value vectors. Between cross-attention layers, we use the residual connection to preserve the original node information. Through successive cross-attention layers, the node embeddings are progressively enriched with lookahead information, yielding updated representations:
\begin{equation}
\label{eq:HPFN_update_node_features}
\mathcal{V}_t^\prime = \text{Softmax}\Bigg( \frac{\mathcal{W}_q\mathcal{V}_t (\mathcal{W}_k\hat{S}_{t+1}^o)^\top}{\sqrt{d}} \Bigg) \mathcal{W}_v\hat{S}_{t+1}^o,
\end{equation}
where $\mathcal{W}_*$ denotes a learnable weight matrix, and the subscripts $q,k$ and $v$ refer, respectively, to the query, key, and value projections in the attention mechanism. $d$ is the scaling factor.

To incorporate this lookahead prediction into the navigation policy, the updated node embeddings $\mathcal{V}_t^\prime$ are fed into the policy head to compute the action-probability distribution for the next time step:
\begin{equation}
\begin{aligned}
    {a}_t^{\prime\prime}&=\mathop{\max}\big(\operatorname{Softmax}(\operatorname{FFN}(\mathcal{V}_t^\prime))\big).
\end{aligned}
\label{eq:HPFN_action2}
\end{equation}

\section{Experiments}
\subsection{Experiment Settings}
\subsubsection{Datasets}
 Our work primarily focus on the path instruction following task on the Room-to-Room (R2R) benchmark dataset ~\cite{anderson2018vision_r2r}. R2R requires an agent to follow step-by-step natural language instructions, averaging 32 words in length. To rigorously evaluate the generalization capabilities of our approach, we additionally conduct evaluations on the REVERIE benchmark ~\cite{qi2020reverie}. REVERIE presents more challenging high-level instructions (averaging 21 words), such as "Find the cell phone on the nightstand in the bedroom", and requires the agent to navigate to the target location and localize a distant target object within a pre-defined ground-truth bounding box. Notably, the expert demonstration path lengths in REVERIE range from 4 to 7 steps, compared to an average of 6 steps in R2R.

\subsubsection{Evaluation Metrics}
We adopt the widely recognized evaluation framework in vision-and-language navigation. For core navigation performance assessment, key metrics include: Success Rate (SR), measuring the proportion of paths where the agent's final position is within a preset threshold (typically 3 meters) from the goal; Success weighted by Path Length (SPL), which jointly considers task success and path efficiency; Trajectory Length (TL) representing the average path length in meters; Navigation Error (NE) indicating the average distance between the agent's final position and the target; and Oracle Success Rate (OSR) showing success rate under ideal stopping policy. For object referral evaluation, following REVERIE \cite{qi2020reverie}, we employ Remote Grounding Success (RGS) to assess the proportion of successfully reaching the correct area and accurately locating the target object, along with its path-length-weighted variant (RGSPL). Among all metrics, except TL and NE where lower values indicate better performance, higher values represent superior results for other indicators.

\subsubsection{Implementation Details}
To demonstrate that UNeMo serves as a generalizable framework applicable to diverse baseline architectures, we validate its effectiveness on two distinct topology-based navigation policies: NavGPT2 \cite{zhou2024navgpt2} and DUET \cite{chen2022think_topo}. Both policies share a common limitation: while capable of processing vision-language inputs and constructing topological representations via imitation learning, they lack prospective state reasoning capabilities. To address this gap, we integrate explicit state reasoning tasks into both systems using the UNeMo framework. Our experiments retain all baseline pretraining results and introduce UNeMo exclusively during the fine-tuning phase. An important implementation distinction lies in their modality processing: DUET employs conventional vision-language pretraining for feature extraction, whereas NavGPT2 acquires representations via LLM intermediate layer distillation. This difference provides a controlled setting for evaluating UNeMo's adaptability to diverse multimodal inputs. All experiments were conducted on Ubuntu 22.04 using NVIDIA RTX 4090 GPUs (single-GPU configuration), with complete environment specifications to be provided in our open-source code repository later.

\begin{table*}[t]
\centering
\small
\begin{tabular}{@{}>{\centering\arraybackslash}p{3.2cm}>{\centering\arraybackslash}p{0.9cm}*{10}{>{\centering\arraybackslash}p{0.78cm}}@{}}
\toprule
\multirow{2}{*}{Methods} & \multirow{2}{*}{\makecell{Freeze\\LLM}} & \multicolumn{5}{c}{Val Unseen} & \multicolumn{5}{c}{Test Unseen} \\
\cmidrule(lr){3-7} \cmidrule(l){8-12} & & 

TL & NE$\downarrow$ & OSR$\uparrow$ & SR$\uparrow$ & SPL$\uparrow$ \ & TL & NE$\downarrow$ & OSR$\uparrow$ & SR$\uparrow$ & SPL$\uparrow$ \\ \midrule
NavCoT(LLaMa2-7B) & No & 9.9 & 6.3 & 48 & 40 & 37 & - & - & - & - & - \\
LangNav(LLaMa2-7B) & No & - & - & - & 46 & - & - & - & - & - & - \\
NaviLLM(Vicuna-7B) & No & 12.8 & 3.5 & - & 67 & 59 & 13.2 & 3.7 & - & 68 & \underline{60} \\
\midrule
NavGPT2(FlanT5-1.5B) & Yes & 12.8 & 3.3 & 79 & 70 & 59 & 13.5 & 3.4 & 77 & \underline{71} & \underline{60} \\
NavGPT2(FlanT5-5B) & Yes & 13.6 & \underline{3.1} & \textbf{81} & \underline{72} & \underline{61} & 14.6 & \underline{3.4} & \underline{78} & \underline{71} & \underline{60} \\
\midrule
UNeMo(FlanT5-1.5B) & Yes & 13.1 & \textbf{3.0} & \underline{79.6} & \textbf{72.1} & \textbf{61.1} & 13.9 & \textbf{3.1} & \textbf{78.4} & \textbf{72.5} & \textbf{61.3} \\

\bottomrule
\end{tabular}
\caption{Performance comparison with SOTA LLM-based VLN methods on R2R.}
\label{tab:R2R_performance}
\end{table*}

\begin{center}
\begin{table*}[t]
\centering
\small
\begin{tabular}{@{}>{\centering\arraybackslash}p{1.8cm} 
                  >{\centering\arraybackslash}p{0.7cm}  
                  >{\centering\arraybackslash}p{0.8cm}  
                  >{\centering\arraybackslash}p{0.7cm}  
                  >{\centering\arraybackslash}p{0.8cm}  
                  >{\centering\arraybackslash}p{0.8cm}  
                  >{\centering\arraybackslash}p{1.2cm}  
                  >{\centering\arraybackslash}p{0.7cm}  
                  >{\centering\arraybackslash}p{0.8cm}  
                  >{\centering\arraybackslash}p{0.7cm}  
                  >{\centering\arraybackslash}p{0.8cm}  
                  >{\centering\arraybackslash}p{0.8cm}  
                  >{\centering\arraybackslash}p{1.2cm}@{} 
                  }
\toprule
\multirow{2}{*}{{Methods}} & \multicolumn{6}{c}{{Val Unseen}} & \multicolumn{6}{c}{{Test Unseen}} \\
\cmidrule(lr){2-7} \cmidrule(l){8-13} 
 & {TL} & {OSR$\uparrow$} & {SR$\uparrow$} & {SPL$\uparrow$} & {RGS$\uparrow$} & {RGSPL$\uparrow$} & {TL} & {OSR$\uparrow$} & {SR$\uparrow$} & {SPL$\uparrow$} & {RGS$\uparrow$} & {RGSPL$\uparrow$} \\
\midrule
RecBERT & 16.78 & 35.02 & 30.67 & 24.9 & 18.77 & 15.27 & 15.86 & 32.91 & 29.61 & 23.99 & 16.5 & 13.51 \\
AirBert & 18.71 & 34.51 & 27.89 & 21.88 & 18.23 & 14.18 & 17.91 & 34.2 & 30.28 & 23.61 & 16.83 & 13.28 \\
HAMT & 14.08 & 36.84 & 32.95 & 30.2 & 18.92 & 17.28 & 13.26 & 33.41 & 30.40 & 26.67 & 14.88 & 13.08 \\
DUET* &26.17 & \underline{51.95} & 46.66 & 31.03 & \underline{32.32} & 21.41 & 24.25 & \underline{57.91} & \underline{52.61} & 34.82 & 31.12 &20.71 \\
DUET &22.11 &51.07 &\underline{46.98} &\textbf{33.73} & 32.15 &\textbf{23.03} &{21.30} &56.91 &52.51 &\textbf{36.06} &\underline{31.88} &\textbf{22.06} \\
UNeMo &24.46 &\textbf{53.99} &\textbf{49.36} &\underline{32.84} &\textbf{34.28} &\underline{22.90} &{23.90} &\textbf{58.82} &\textbf{53.21} &\underline{35.17} &\textbf{32.15} &\underline{20.90} \\
\bottomrule
\end{tabular}
\caption{Performance Comparison on REVERIE Dataset. *indicates reproduced results.}
\label{tab:REVERIE_performance}
\end{table*}
\end{center}

\subsection{Comparison with State of the Art}
\subsubsection{R2R Result}
For the R2R dataset experiments, we compared our method with state-of-the-art (SOTA) approaches in LLM-based vision-and-language navigation, including NavCoT \cite{lin2025navcot}, LangNav\cite{pan2023langnav}, NaviLLM\cite{zheng2024NaviLLM}, and NavGPT2\cite{zhou2024navgpt2}. These methods formulate navigation as a language modeling problem, employing either prompt engineering or simple imitation learning paradigms for end-to-end navigation decision-making.

As shown in Table 1, the results demonstrate that our method outperforms several competitive methods on both val unseen and test unseen splits. Specifically, on the val unseen split, our approach achieves improvements of 2.1\% in SR and SPL compared to baseline methods. For the more challenging test unseen split, these key metrics show gains of 1.5\% and 1.3\%, respectively. Notably, UNeMo exhibits superior efficiency: with only a FlanT5-1.5B backbone (30\% parameters of NavGPT2's FlanT5-5B), it requires less than half the GPU memory (12GB vs 27GB) while delivering better navigation performance. Furthermore, UNeMo demonstrates particular robustness in long-horizon navigation: on the R2R val-seen split, it exhibits significantly larger SR gains on long paths than on short paths. For long paths with a length of $\geq$7, its SR increases from 64.2\% to 69.8\%, a 5.6\% improvement, whereas for short paths with a length of $<$7, the SR only rises slightly from 71.1\% to 72.3\%, with a gain of 1.2\%. This indicates its enhanced capability in handling complex, long-trajectory navigation scenarios.

These results confirm two core advantages of our method: (1) high efficiency via a compact model that reduces GPU memory footprint while retaining strong performance; (2) superior long-horizon robustness from the multimodal framework’s enhanced reasoning. This provides a practical path for developing efficient, scalable LLM-based navigation systems.

\subsubsection{REVERIE Results}
We validate UNeMo’s effectiveness on goal-oriented navigation via the REVERIE dataset, with comparative experimental results presented in Table ~\ref{tab:REVERIE_performance}. DUET denotes the original results from the method’s paper, while DUET* represents our reproduced results—and UNeMo outperforms DUET* across all metrics. Specifically, UNeMo achieves a navigation SR of 49.36\% on Val Unseen and 53.21\% on Test Unseen, surpassing both DUET and DUET*. Its remote RGS reaches 34.28\% on Val and 32.15\% on Test, outperforming DUET’s 32.15\% and 31.88\% as well as DUET*.
Notably, UNeMo shows marginal drops in path efficiency metrics SPL and RGSPL. This stems from REVERIE’s natural language instructions providing only coarse-grained object descriptions, unlike R2R’s detailed route guidance, which makes REVERIE’s task emphasize agents’ exploration and path correction capabilities. Unlike DUET’s sole reliance on imitation learning, UNeMo’s hierarchical prediction-feedback mechanism incorporates prospective exploration before navigation decisions, enabling course correction for excessive deviations. This may require extra exploratory actions but ensures target arrival. The trade-off between path efficiency and task success aligns with practical goal-oriented navigation needs, confirming UNeMo prioritizes core navigation requirements—success rates—over marginal efficiency losses.

\begin{table*}[htbp]
\centering 
\begin{tabular}{@{} cccccccc@{}}
\toprule \multirow{2}{*}{Methods} & \multicolumn{2}{c}{Modality} & \multicolumn{5}{c}{Val Unseen} \\
\cmidrule(lr){2-3} \cmidrule(lr){4-8} & Vision & Language & TL & NE$\downarrow$ & OSR$\uparrow$ & SR$\uparrow$ & SPL$\uparrow$ \\
\midrule
NavGPT2 & \texttimes & \texttimes & 12.8 & 3.33 & 79 & 70 & 59 \\
\midrule
TopoState & \texttimes & \texttimes & 13.2 & 3.08 & 80 & 72.2 & 60.7 \\
Cond2Vis & \checkmark & \texttimes & 12.6 & 3.09 & 79.7 & 72.4 & 61.7\\
\midrule
Vis-WM & \checkmark & \texttimes & 13.4 & 3.18 & \textbf{80.6} & 71.8 & 60.7 \\
\textbf{UNeMo} & \checkmark & \checkmark & 13.3& \textbf{3}& \textbf{80.6}& \textbf{72.9}& \textbf{61.7}\\
\bottomrule \end{tabular} 
\caption{Comparison of Different State Reasoning Methods. ``TopoState" denotes the topology-map-based approach, ``Cond2Vis" denotes vision-based approach, ``Vis-WM" denotes the vision-based world model, ``UNeMo" denotes our multi-modal-based world model.}
\label{tab:ablation_study_of_state_inference}
\end{table*}

\subsection{Ablation Study}
This section presents systematic ablation experiments to validate the effectiveness of different state reasoning methods and hierarchical prediction feedback mechanisms.

\subsubsection{Comparison of Different State Reasoning Methods}
Since state reasoning’s effectiveness hinges on the chosen approach, we develop several variants to identify the most suitable method. We propose a state deduction framework that enhances multi-modal reasoning through future state prediction, implemented via two paradigms: topology-map-based state prediction and vision-based state decoding.

In the topology-map-based approach, the agent dynamically constructs a graph-structured topological map $\mathcal{G}_t$ where node features are aggregated via average pooling to form the current state representation $\mathcal{S}_t$. This representation is transformed through an MLP, then updated to $\mathcal{S}_{t+1}$ after expert action execution. As shown in Row 2 (TopoState) of Table \ref{tab:ablation_study_of_state_inference}, this method achieves 2.2\% and 1.7\% improvements in SR and SPL, respectively. However, it suffers from label acquisition constraints—requiring additional navigation actions and map updates to obtain state labels.

\begin{table}[t]
\centering
\begin{tabular}{@{}c cc ccc@{}}
\toprule \multirow{2}{*}{Methods} & \multicolumn{5}{c}{Val Unseen} \\
 \cmidrule(l){2-6}
& TL & NE$\downarrow$ & OSR$\uparrow$ & SR$\uparrow$ & SPL$\uparrow$ \\
\midrule
MWM only & 13.1 & 3.3 & 79.7 & 71.5 & 61.0 \\
\midrule
$a^\prime$ only & 13.1 & {3.0} & 79.6 & 72.1 & 61.1 \\
$a^{\prime\prime}$ only & {13.3} & \textbf{3.0} & \textbf{80.6} & \textbf{72.9} & \textbf{61.7} \\
$a^\prime$ and $a^{\prime\prime}$& 14.1 & 3.2 & 80.3 & 72.1 & 60.3 \\
\bottomrule
\end{tabular}
\caption{Ablation Study on Dual-Phase Action Learning policies in Hierarchical Prediction-Feedback Navigator.}
\label{tab:ablation_HPFN}
\end{table}

We provide a vision-based alternative: directly predicting candidate view features of expert-action-targeted nodes as future state representations. Row 3 (Cond2Vis) of Table \ref{tab:ablation_study_of_state_inference} demonstrates 2.4\% SR and 2.7\% SPL gains on R2R val-unseen data. Notably, predicted states do not directly influence decisions but implicitly optimize node encoding features $\mathcal{V}_t$ through gradient backpropagation. Both methods significantly improve SR and SPL metrics, proving that this design enables node embedding spaces to autonomously embed latent environmental state dynamics, thereby enhancing navigation performance.

We also conduct comprehensive ablation studies on the R2R dataset's val unseen split to evaluate our UNeMo framework's Multi-modal World Model. As demonstrated in Table \ref{tab:ablation_study_of_state_inference}, the experimental results reveal that utilizing local observations of future states from current nodes as input, coupled with decoding compressed visual features through the joint navigation module's condition features (Cond2Vis), significantly enhances the agent's navigation capability (SR: 72.4\% vs. baseline 70\%). However, the world model-based visual feature decoding approach (Vis-WM), which predicts future states solely from partial observations, shows marginally inferior performance (SR: 71.8\%) compared to Cond2Vis. To address this limitation, we incorporate natural language instructions containing action prompts and scene descriptions, implementing cross-attention fusion with local observations of visual features (UNeMo). The final results achieve state-of-the-art (SOTA) performance, exhibiting 2.9\% (SR: 72.9\%) and 2.7\% (SPL: 61.7\%) improvements over the baseline.

In conclusion, our ablation experiments substantiate the efficacy of state reasoning methods for LLM-based Vision-and-Language Navigation (VLN) systems. Considering the practical challenges in acquiring topological map state labels, we propose a novel approach that leverages Multimodal World Model for vision-language data fusion to infer future visual states, while systematically validating the utility of each modular component.

\subsubsection{Ablation Analysis of Hierarchical Prediction-Feedback Navigator}
To rigorously validate the pivotal role of the hierarchical prediction-feedback mechanism within the UNeMo framework, this study constructs a dedicated baseline model that explicitly excludes the Hierarchical Prediction-Feedback Navigator (HPFN) for systematic comparative analysis. As demonstrated in Table \ref{tab:ablation_HPFN}, when the framework solely employs the MWM as an auxiliary training branch without the integration of HPFN, the navigation metrics—including key indicators such as success rate, path length, and normalized dynamic time warping—on the validation Unseen set exhibit a substantial decline, rendering them significantly inferior to those of the optimized solution that incorporates HPFN. This compelling empirical evidence directly substantiates the critical function of the hierarchical decision-making mechanism in enhancing overall navigation performance, as it bridges the gap between visual state reasoning and actionable navigation policies.

Furthermore, to gain deeper insights into the optimal configuration of HPFN, we designed comparative experiments to evaluate different learning policies for HPFN's dual-node scoring mechanism, systematically examining three distinct approaches: learning solely from action1 scoring($a^\prime$ only), learning solely from action2 scoring($a^{\prime\prime}$ only), and joint learning from both scoring processes($a^\prime$ and $a^{\prime\prime}$). Our experimental results demonstrate that the learning policy focusing on the second scoring output - which integrates state inference results from MWM - optimally balances navigation accuracy and path efficiency in HPFN, thereby establishing the most effective decision-making framework for the UNeMo method.

\section{Conclusion}
This paper addresses the key challenges in Vision-and-Language Navigation (VLN), where existing LLM-based methods are limited by insufficient visual reasoning capabilities and disjoint optimization between reasoning modules and navigation policies. We propose UNeMo, a novel framework that enables collaborative optimization of visual state reasoning and navigational decision-making through its Multimodal World Model (MWM) and hierarchical mechanism. MWM facilitates cross-modal reasoning by jointly predicting subsequent visual states from visual features, language instructions, and navigational actions. The hierarchical interaction between MWM and navigation policies establishes a dynamic bidirectional promotion mechanism, where MWM reasoning enhances policy optimization while policy decisions refine MWM’s reasoning accuracy. Experimental results on R2R and REVERIE datasets demonstrate that UNeMo outperforms state-of-the-art methods, validating its effectiveness in improving VLN performance.
This work highlights the potential of integrating multimodal world modeling with hierarchical decision-making mechanism, offering a new direction for developing more robust and adaptive VLN agents. 
Furthermore, benefiting from Matterport3D's real-world data, UNeMo narrows the visual gap between simulation and reality.
Our future work will extend UNeMo to more complex settings and validate its real-world navigation performance via physical robot deployment.

\section{Acknowledgements}
This work is supported in part by the National Natural Science Funds for Distinguished Young Scholar under Grant 62325307, in part by the National Natural Science Foundation of China under Grants (6240020443, 62073225, 62203134, 62373258), in part by the Natural Science Foundation of Guangdong Province under Grants 2023B1515120038, in part by Shenzhen Science and Technology Innovation Commission (20231122104038002, 20220809141216003, KJZD20230923113801004, KJZD20230923115215032, JCYJ20240813141628038), in part by the Guangdong “Pearl River Talent Recruitment Program” under Grant 2019ZT08X603, in part by the Guangdong “Pearl River Talent Plan” under Grant 2019JC01X235, in part by the Scientific Instrument Developing Project of Shenzhen University under Grant 2023YQ019, in part by the Foundation for Distinguished Young Talents in Higher Education of Guangdong under Grant 2025KQNCX112.

\section{Appendix}

\subsection{Implementation Details of Auto-Encoder Training} \label{sec:autoencoder}
To avoid predicting complex and dense visual representations, a module is required to compress them.
To enhance the adaptability of the visual compression model to the Visual-Language Navigation (VLN) task, a dedicated training set is constructed based on the R2R dataset: 30,852 views oriented toward candidate nodes are selected from 207,380 panoramic views, and the original visual features $E^o \in \mathbb{R}^{1 \times H \times W}$ extracted by a frozen Vision Transformer (ViT) are used as inputs. The visual compression module adopts an encoder-decoder architecture: the encoder compresses features into 768-dimensional latent codes through three-level convolution (with channel numbers 1→32→64→128) and a Multi-Layer Perceptron (MLP); the decoder reconstructs the input features via transposed convolution and MLP. During training, the information density of latent codes is optimized with the goal of minimizing the mean squared error loss. In the fine-tuning phase, the feature of N views (with a shape of $N\times1\times H\times W$) at node j is input into the compression module in batches. The output, after average pooling, serves as the visual state representation of the node, denoted as $S^o \in \mathbb{R}^{1 \times 768}$.

\subsection{Exploratory Study on State Reasoning Methods}
Prior to finalizing the multimodal world model as our ultimate visual state reasoning approach, we systematically investigated various alternative solutions (as detailed in the ablation studies section). These exploratory endeavors laid crucial groundwork for the development of the UNeMo method and provided significant insights into the pivotal role of state reasoning in agent navigation. This section elaborates on the design and implementation of these methods, followed by a comparative analysis of their performance on the R2R dataset \cite{anderson2018vision_r2r}.
\subsubsection{Topology map-based state reasoning approach}
Inspired by the topological mapping mechanism in NavGPT2 \cite{zhou2024navgpt2}, we identified two critical phases in its navigation pipeline: (1) updating node encodings in the topological map based on current visual observations and natural language instructions upon arriving at each node, and (2) scoring globally navigable nodes according to the node encodings in the topological map (see the Preliminary section for details). Building upon this framework, we designed a supervised learning task to enhance the agent's state reasoning capability. Specifically, we first aggregate all node features $\mathcal{V}_t = \{ \mathcal{V}_{t,i} \}_{i=1}^K$ 
in the current topological map $\mathcal{G}_t$ (where $K$ denotes the number of nodes), 
via average pooling to obtain the current state representation $S^{m}_{t}$: 
\begin{equation}
S^{m}_{t} = \frac{1}{N} \sum_{i=1}^{N} \mathcal{V}_{t,i}, 
\end{equation}
Then, using a multilayer perceptron-based decoder, we enable the agent to predict the next state representation $S^{m}_{t+1}$ after executing the ground truth navigation action $a_{gt}$, based on $S^m_{t+1}$.

\subsubsection{Vision-Based Decoder Schemes}
While the topology-based visual state construction proves effective, it suffers from the difficulty in acquiring next-state topology labels—obtaining $S^m_{t+1}$ necessitates complete navigation execution and topological graph updates, significantly increasing training duration and computational overhead. Furthermore, existing visual fusion mechanisms in LLM-based VLN methods exhibit limitations: either simplifying fine-grained visual information into generalized textual descriptions or directly feeding raw visual tokens into language models, both leading to critical visual information loss. To enhance scene change perception and visual comprehension, we propose two prediction schemes leveraging topological node encodings:

\subsubsection{Dense Visual Prediction}
Utilizes the current topological node encoding $\mathcal{V}_t$ to predict the dense representation of visual observations at candidate nodes. The target representation is obtained via Vision Transformer (ViT) from next-state visual observations $O_{t+1}$:
\begin{equation}
S^{v}_{t+1} = \text{ViT}(O_{t+1}), \quad \hat{S}^{v}_{t+1} = \text{MLP}(V_{t})
\end{equation}
where $\hat{S}^{o}_{t+1} \in \mathbb{R}^{257 \times 1408}$ denotes the predicted dense representation, $S^{o}_{t+1}$ represents the actual visual embedding acquired after action execution.

\subsubsection{Sparse Visual Prediction}
Employs a Conditional Variational Autoencoder (CVAE) to predict compressed representations of post-action visual observations from partial views $O_t$. First, an autoencoder compresses the full visual representation of ground-truth nodes: 
\begin{equation}
S^{o}_{t+1} = \text{Auto-Encoder}(E^o_{t+1}), 
\end{equation}
where $E^o_{t+1} = Vit(O_{t+1})$, $\text{Auto-Encoder}(\cdot)$ refers to the pre-trained auto-encoder described in Section \ref{sec:autoencoder}. The next-state representation ${S}^{o}_{t+1} \in \mathbb{R}^{1 \times 256}$ reduces dimensionality by three orders of magnitude compared to previous schemes, significantly lowering computational complexity.

\subsubsection{Summary of State Prediction Approaches}
Experimental results demonstrate consistent improvements over baselines across all state prediction schemes. While full-scale visual prediction achieves optimal accuracy, its computational demands exceed RTX 4090's 24GB VRAM capacity when processing topological maps. Compressed feature prediction reduces resource usage but suffers from representational ambiguity. To address these limitations, we propose a Multimodal World Model (MWM) combining both approaches.

\begin{figure*}[t]
    \centering
    \includegraphics[width=0.95\textwidth]{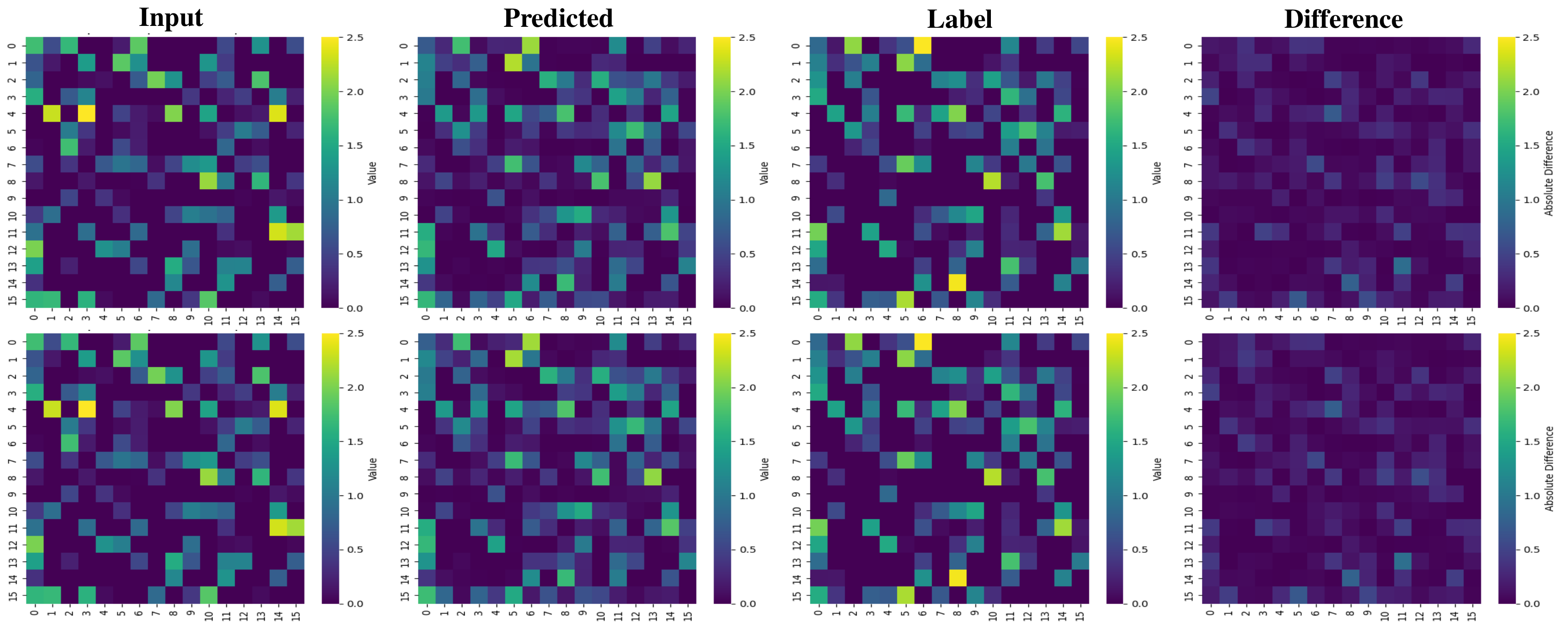}
    \caption{Qualitative comparison between MWM predicted features and ground-truth labels under partial observations}
    \label{fig:feature_compare}
\end{figure*}

\begin{figure*}[!t]  
    \centering
    \includegraphics[width=0.95\textwidth]{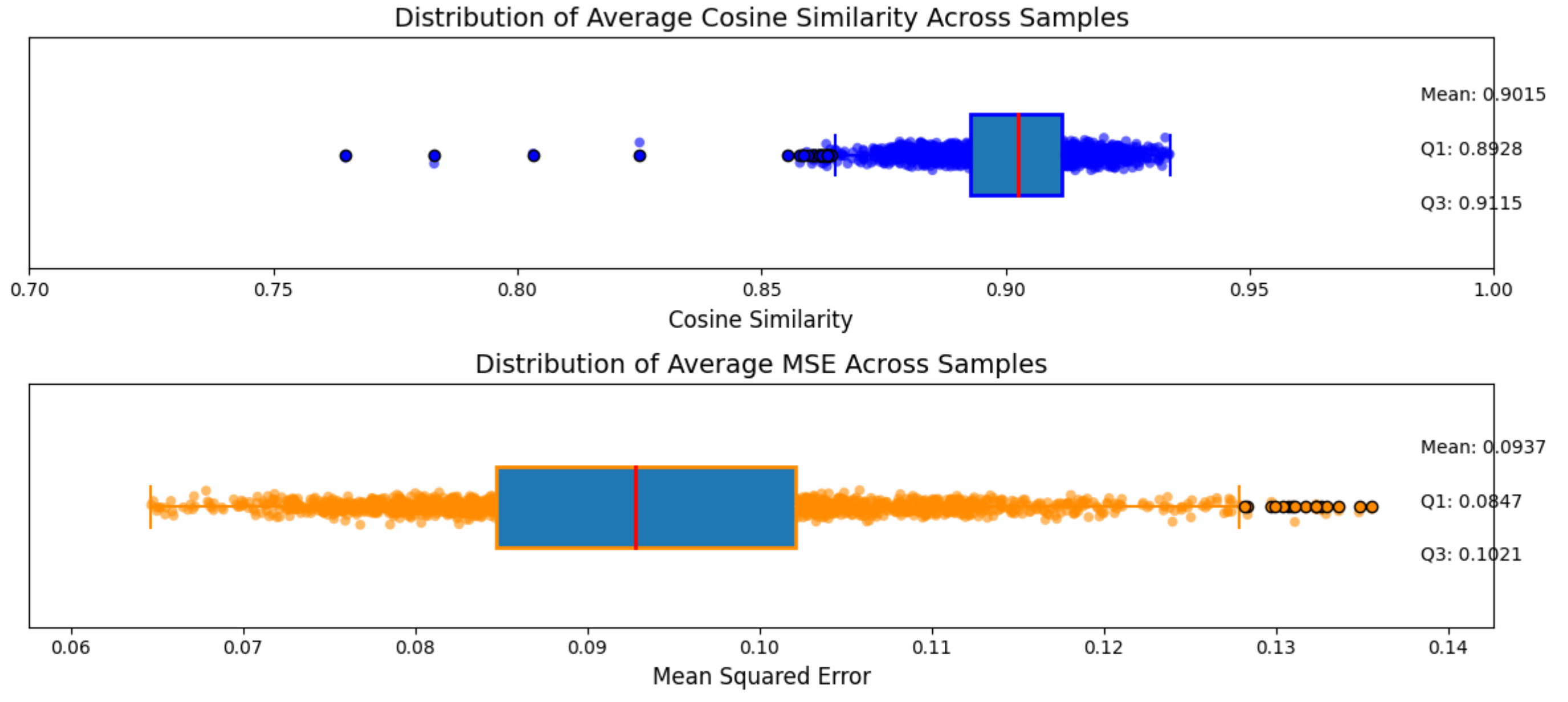}
    \caption{Quantitative evaluation of MWM predictions: Distributions of cosine similarity and MSE against labels in val-unseen}
    \label{fig:status_compare}
\end{figure*}

\subsection{UNeMo Implementation Details}
During fine-tuning, we batch-process N view features (shape N×1×H×W) at node j through the compression module, with the outputs averaged via pooling to form the node's visual state representation $S^{o} \in \mathbb{R}^{1 \times 768}$.

\subsubsection{aseline implementation differences}
The multimodal state reasoning module in the UNeMo framework requires both visual-linguistic modality data as input and incorporates topological map node encodings as conditional information. Experimental validation was conducted on two topology-based baseline methods: the NavGPT2-based UNeMo implementation (detailed in the Methods section) and the DUET-based modified version. The primary distinction lies in their multimodal feature extraction approaches - the former employs frozen latent representations from a large language model (LLM) for visual-linguistic encoding, whereas the latter utilizes representations generated by trainable Transformer blocks.

\begin{table*}[htbp]
\centering
\caption{Performance Comparison on R2R Dataset}
\label{tab:r2r_results}
\begin{tabular}{lcccccc}
\toprule
\multirow{2}{*}{Method} & \multirow{2}{*}{\begin{tabular}{c}Freeze \\ LLM\end{tabular}} & \multicolumn{2}{c}{Val Unseen} & \multicolumn{2}{c}{Test Unseen} \\
\cmidrule(lr){3-4} \cmidrule(lr){5-6}
& & SR$\uparrow$ & SPL$\uparrow$ & SR$\uparrow$ & SPL$\uparrow$ \\
\midrule
NavGPT2 (1.5B) & \checkmark & 70.0 & 59.0 & 71.0 & 60.0 \\
NavGPT2 (5B)   & \checkmark & 72.0 & 61.0 & 71.0 & 60.0 \\
\textbf{UNeMo (1.5B)} & \checkmark & \textbf{72.1} & \textbf{61.1} & \textbf{72.5} & \textbf{61.3} \\
\bottomrule
\end{tabular}
\end{table*}

\begin{table}[h]
\centering
\caption{Generalization Performance on REVERIE Dataset}
\label{tab:reverie_results}
\begin{tabular}{lcccccc}
\toprule
\multirow{2}{*}{Method} & \multicolumn{2}{c}{Val Unseen} & \multicolumn{2}{c}{Test Unseen} \\
\cmidrule(lr){2-3} \cmidrule(lr){4-5}
& SR$\uparrow$ & SPL$\uparrow$ & SR$\uparrow$ & SPL$\uparrow$ \\
\midrule
DUET* & 46.66 & 31.03 & 52.61 & 34.82 \\
\textbf{UNeMo} & \textbf{49.36} & \textbf{32.84} & \textbf{53.21} & \textbf{35.17} \\
\bottomrule
\end{tabular}
\end{table}

\subsubsection{Training strategy details}
To prevent overfitting in the multimodal state prediction module and enhance multi-task training stability during navigation tasks, we adopt a phased training strategy. Specifically, the training process is divided into multiple phases where state reasoning tasks are only activated during the initial 10\% of training batches in each phase. This design addresses two critical considerations: first, while multimodal state reasoning primarily aims to improve navigation performance by enhancing the agent's understanding of sequential navigation processes, direct joint training would cause premature convergence of the CVAE module due to insufficient state reasoning data, thereby compromising the learning efficacy of the navigation policy network; second, since conditional information dynamically evolves with the training of the navigation policy network, we periodically retrain the Multimodal World Model (MWM) to maintain alignment with the shifting feature space.

All experiments in this section were conducted on Ubuntu 22.04 LTS, with the R2R and REVERIE datasets running on NVIDIA RTX 4090 and RTX 5090 GPUs respectively, both supporting single-GPU testing. Detailed software dependencies and third-party library configurations can be found in the requirements list of the open-source code.

\subsection{Visualization Analysis of State Prediction Results}
To elucidate the principles underlying the effectiveness of our method, we conducted a visual analysis of the state inference results. We randomly sampled timesteps from all trajectories in the val unseen validation set scenarios and visualized the inference outcomes of the MWM world model, as illustrated in Figure \ref{fig:feature_compare}. The visualization of randomly sampled state features in the first and second rows of the figure demonstrates that, given different local views as input, the world model can infer results closely approximating the next-state labels, with the predicted features exhibiting distributional characteristics similar to those of the label features.

Furthermore, we performed a statistical analysis of the inference results across all samples in the val unseen scenarios, computing both the cosine similarity and mean squared error between the predicted representations and the state labels. The detailed visualization of these metrics is presented in Figure \ref{fig:status_compare}. The results reveal that the average cosine similarity between our predicted next-state representations and the ground-truth labels approaches 90.15\%, with an average MSE of approximately 0.09. These findings robustly demonstrate the capability of our inference module to perform accurate prospective exploration in unseen scenarios.

\bibliography{aaai2026}
\end{document}